\documentclass[11pt]{article}

\usepackage[utf8]{inputenc}
\usepackage[T1]{fontenc}
\usepackage{lmodern}
\usepackage{microtype}
\usepackage{geometry}
\usepackage{graphicx}
\usepackage{booktabs}
\usepackage{tabularx}
\usepackage{multirow}
\usepackage{array}
\usepackage{enumitem}
\usepackage{xcolor}
\usepackage{amsmath}
\usepackage{amssymb}
\usepackage{url}
\usepackage[hidelinks]{hyperref}
\usepackage[round,authoryear]{natbib}
\usepackage{caption}
\usepackage{subcaption}
\usepackage{float}

\graphicspath{{figures/}}
\setlist{nosep}
\newcommand{\system}{\textsc{Aegix Pulse}}

\title{\system: A Traceable Three-Stage Architecture for Personalized Content Generation and Context-Preserving Revision}

\author{
Hongnan Zhao$^{1}$ \quad
Shiyu Chen$^{1,*}$ \quad
Zhihao Chen$^{1}$\\
{\small $^{1}$Aegix Insight}\\
{\small $^{*}$Corresponding author: \texttt{shannon.chen@aegix.top}}\\
{\small Emails: \texttt{evan.zhao@aegix.top},
\texttt{shannon.chen@aegix.top},
\texttt{zhihao.chen@aegix.top}}
}

\date{}

\begin{document}
\maketitle

\begin{abstract}
Production content-generation systems must bring together a user's immediate task, a brand's long-term identity, evidence from successful historical content, and feedback provided during later revisions. However, adding more context does not necessarily lead to better content, and existing system descriptions rarely measure the contribution of each context layer separately. We present \system, a production-oriented three-stage architecture that separates current-task clarification and Task Persona finalization, long-term Account Profile (Brand DNA) assembly, and controlled generation and revision, while preserving immutable provenance across content versions. 

We evaluate four focused claims in a preregistered offline experiment involving 96 synthetic social-media generation tasks. Four initial-generation conditions (E0--E3) progressively introduced a Task Persona, an Account Profile, and successful-history style evidence. Two revision conditions (R0--R1) compared plain revision with context-preserving revision. The experiment produced 480 completed generation records and 1,440 blinded LLM-Judge evaluations, together with a stratified human review and adjudication by a third reviewer.

Both the Account Profile and context-preserving revision conditions achieved higher mean scores than their respective baselines, providing encouraging preliminary evidence of improvement. Adding the Account Profile increased mean brand-consistency scores by 0.1562 points on a five-point scale compared with using the Task Persona alone (Holm-adjusted $p=.1224$). Providing the preserved task and brand context during revision increased mean task-preservation scores by 0.2917 points compared with plain revision (Holm-adjusted $p=.2432$). Although the current evidence was not sufficient to establish these improvements as statistically conclusive after multiple-comparison correction, the results suggest that persistent brand context and preserved task context during revision may improve personalized content generation.

Task Persona alone produced a small observed effect, while successful-history style evidence did not provide an additional improvement in brand consistency beyond Task Persona plus Account Profile under the current experimental setting. This result does not imply that historical evidence is inherently ineffective. Instead, it identifies several areas for further improvement, including evidence qualification, extraction quality, relevance weighting, and the resolution of conflicts between historical patterns and current task requirements.

Human validation did not consistently reproduce the directions of the effects measured by the LLM Judge, and agreement between the original human reviewers was low. We therefore plan to strengthen the evaluation protocol through clearer scoring criteria, reviewer calibration, and validation by additional independent evaluator groups. Taken together, these findings provide concrete directions for improving both the system and its evaluation, rather than a basis for concluding that the corresponding components have no value.
\end{abstract}


\section{Introduction}

Large language models can generate fluent social-media content from a short instruction, but production use introduces requirements that are not captured by fluency alone. A creator may have an immediate content goal, a persistent brand position, prior posts whose expression patterns have performed well, source material that constrains factual claims, and subsequent feedback that requests a narrow change without redefining the task. Treating all of these inputs as one undifferentiated prompt makes it difficult to decide which evidence should dominate, to preserve state across revisions, or to explain why a particular output was produced.

Personalized language generation research has shown that user histories can support tailored outputs, including through retrieval over profile items. The LaMP benchmark formalized multiple personalized classification and generation tasks and demonstrated the value of retrieval-augmented personalization \citep{salemi2024lamp}. Work on retrieval-augmented generation separates parametric generation from external non-parametric evidence \citep{lewis2020rag}, while long-term-memory systems such as MemGPT organize information across memory tiers to support multi-session interaction \citep{packer2023memgpt}. These lines of work motivate explicit external context, but they do not answer a practical architectural question: should the current task, persistent account identity, and historically successful expression be represented as the same object?

\system{} answers this question by separating them. A \emph{Task Persona} is an immutable specification of the current content task. An \emph{Account Profile}, referred to in product terms as the user's \emph{Brand DNA}, is a versioned, longer-term representation of brand identity. \emph{Retrieved Memories} can additionally provide successful-history style evidence: expression patterns supported by historical content that satisfies a versioned success policy. This evidence supplements rather than redefines the Account Profile. These objects remain conceptually and operationally distinct as they flow into Stage 3 generation. Revision is also separated from task redefinition: local feedback stays within a versioned Stage 3 revision chain, whereas changes to audience, topic, core problem, value proposition, content goal, or audience stage require a new Stage 1 clarification and Task Persona finalization cycle.

This separation improves traceability and makes it possible to measure how much each context layer contributes to output quality. The value of additional context depends on its relevance, consistency, evidence quality, and effective use by the generation model. We therefore evaluate \system{} by adding one context component at a time and comparing each condition with the immediately preceding condition, rather than relying only on a comparison between the complete system and a simple baseline. The study examines four questions: whether the Task Persona improves alignment with the user's current task; whether adding the Account Profile improves brand consistency; whether successful-history style evidence provides a further improvement in brand consistency; and whether preserving the original task and brand context during revision improves task preservation. E0--E3 refer to four initial-generation conditions that add these context components step by step. They should not be confused with the three stages of the production system.

The paper makes three contributions, supported by two types of evidence. The first two contributions concern the system architecture and the research infrastructure. They are supported by implementation records, automated contract tests, production acceptance checks, and audits of the experimental artifacts. The third contribution concerns the measured effects of individual system components and is supported by the preregistered statistical experiment:
\begin{enumerate}
  \item We present a production-oriented three-stage architecture that keeps the user's current task, long-term Brand DNA, evidence from successful historical content, controlled generation, validation, and revision history clearly separated and traceable.
  \item We present a reproducible evaluation workflow for studying individual components of a continuously evolving production system. Before running the formal experiment, we fixed the dataset, experimental conditions, evaluation rules, and stopping criteria. During execution, we preserved every attempt, recorded output hashes, conducted repeated blind LLM evaluations, and added independent human validation.
  \item We report both the observed improvements and the limits of the current statistical evidence without changing or expanding the claims after seeing the results. Account Profile and context-preserving revision produced encouraging preliminary improvements, while successful-history style evidence did not provide an additional improvement in LLM-Judge ratings under the current experimental setting.
\end{enumerate}

This study evaluates offline content quality using controlled synthetic tasks. It does not evaluate real-world exposure, engagement, conversion, revenue, user retention, or long-term user satisfaction, and therefore makes no claims about improvements in these outcomes. Recipe-based structural planning and Validator/Repair are implemented components of the production architecture, but they were not enabled in the formal experiment and are not included in the paper's effectiveness claims. In this paper, ``implemented'' means that a capability exists in the versioned production system and has passed the relevant engineering checks. It does not mean that the capability has already been shown to improve content quality or business outcomes.

\section{Related Work}

\subsection{Persona and personalized language generation}

Persona-based generation has been widely studied in dialogue systems and role-conditioned language models. Recent research, however, distinguishes between role-playing, in which a model adopts a fictional or predefined character, and personalization, in which a model adapts its output using information about a specific user. \citet{tseng2024persona} describe these as related but different uses of persona in LLM systems. PersonaLLM shows that language models can express personality traits specified in a prompt, while also emphasizing the need to evaluate whether those traits are represented accurately and consistently \citep{jiang2024personallm}. In \system{}, the term ``persona'' has a narrower, task-oriented meaning. A Task Persona describes the target audience, core problem, topic, value proposition, content goal, and audience stage for a specific content task; it does not represent a fictional character or a simulated human personality.

LaMP provides a relevant benchmark for studying the use of long-term user information in personalized language generation. Its tasks retrieve items from a user's profile and use them to support personalized text generation \citep{salemi2024lamp}. \system{} makes a further distinction between three types of information: the intent of the current task, the versioned Account Profile (Brand DNA), and style evidence extracted from historical content that meets predefined success criteria. This distinction directly shapes our step-by-step experimental design. E1 uses only the Task Persona, E2 adds the Account Profile, and E3 additionally includes successful-history style evidence.

\subsection{Retrieval and long-term memory}

Retrieval-augmented generation combines knowledge stored in a language model with information retrieved from external sources. This approach allows external knowledge to be updated without retraining the model and can make the information used during generation easier to trace \citep{lewis2020rag}. Research on agent memory applies the same general idea to longer interactions. For example, MemGPT uses a hierarchical memory system to move information between the model's limited active context and external storage \citep{packer2023memgpt}.

\system{} also uses external retrieval, but it gives different storage systems clearly defined responsibilities. Relational records are the authoritative source for auditable entities, versions, and relationships. Vector indexes are used for semantic retrieval and can be rebuilt from the authoritative records when necessary. Historical content becomes eligible for style-evidence retrieval only after it meets a versioned success criterion. The resulting style evidence is stored separately from both the immutable Task Persona and the versioned Account Profile.

In this study, the context provided to the generation model was fixed before the formal experiment and manually checked for accuracy. The experiment therefore measures how each type of context affects output quality when the intended information is supplied correctly. It does not evaluate whether the production retrieval and extraction components can always select the most relevant historical content or extract the correct style evidence.

\subsection{Iterative generation, validation, and revision}

Iterative refinement allows a model to review an initial output, identify problems, and produce an improved version. Self-Refine showed that repeated feedback and revision can improve performance across a range of tasks without requiring additional supervised training \citep{madaan2023selfrefine}.

\system{} applies this idea within a structured production workflow. Recipe-based planning provides reusable requirements for organizing content, while preserving the current task definition and brand context. After generation, the Validator checks brand requirements, factual boundaries, formatting rules, Recipe requirements, and the constraints that define how the two controlled variants may differ. When a violation is detected, the Repair process can revise the output within a limited number of attempts. User-requested revisions follow an explicit contract that records what should remain unchanged and what may be modified. Each revised version is also linked to its immediate parent, creating a traceable version history.

The formal revision experiment is designed to measure the effect of preserving the original task and brand context during revision. R0 and R1 use the same E3-generated source content and receive the same user feedback. R0 revises the content using only the source content and feedback, whereas R1 also receives the finalized Task Persona, Account Profile, successful-history style evidence, and original generation context. Recipe and Validator/Repair are disabled in both conditions. This design ensures that any observed difference between R0 and R1 is associated with the additional preserved context rather than Recipe-based planning or automated repair.

\subsection{Evaluation of open-ended generation}

Traditional reference-based metrics compare generated text with one or more predefined answers. They are often unsuitable for open-ended content-generation tasks because many different outputs may be equally valid. G-Eval showed that an LLM using explicit evaluation criteria can produce judgments that align more closely with human ratings than traditional automatic metrics. It also noted that LLM evaluators may favor text generated by language models \citep{liu2023geval}. MT-Bench and Chatbot Arena further demonstrated that LLM-as-a-Judge can support evaluation at scale, while documenting several important sources of bias, including candidate order, response length, and a model's tendency to favor outputs similar to its own \citep{zheng2023judging}.

To reduce these risks, \system{} fixes the Judge prompt before the formal experiment, hides the experimental condition labels, randomizes the order of the candidate outputs, and evaluates each candidate three times independently. The repeated scores are combined using rules defined before the results are examined. The evaluation also includes independent human review, so LLM-Judge scores are not treated as unquestionable ground truth. In our results, the LLM Judge was relatively consistent across repeated evaluations, but its effect directions did not always agree with those observed in the human-reviewed subset. This disagreement shows that stable LLM ratings do not necessarily mean that the ratings reflect human judgment or measure the intended quality dimensions accurately.

\subsection{Preregistration for evolving AI systems}

Continuously improving an AI system creates a risk of experimental overfitting. After observing an unfavorable result, researchers may repeatedly change the system, dataset, evaluation criteria, or exclusion rules until the experiment produces a more favorable outcome. Preregistration reduces this risk by requiring the hypotheses, experimental design, evaluation rules, and analysis plan to be specified before the formal results are examined \citep{nosek2018preregistration}. This makes it easier to distinguish ideas developed during exploration from evidence produced by a confirmatory experiment.

In the \system{} study, all previously examined Pilot and development-validation cases are treated only as diagnostic data. Before formal execution, we fixed the formal dataset, evaluator, scoring rules, and stopping criteria. We also limited the number of permitted system modifications and retained all failed attempts rather than deleting or replacing them. Once the formal dataset had been revealed, it could no longer be used to tune a later system version and then be presented again as independent confirmatory evidence.

\section{The \system{} Architecture}

\subsection{Design principles}

\system{} is designed around five principles:
\begin{itemize}
  \item \textbf{Keep information with different lifetimes separate.}
  The user's current task, long-term Account Profile, and evidence from successful historical content are stored as separate types of state. This prevents information from one content task from being confused with long-term brand information or historical evidence.
  \item \textbf{Separate authoritative records from semantic retrieval.}
  Relational records are the authoritative source for users, versions, relationships, and audit history. Vector indexes support semantic retrieval and can be rebuilt from the authoritative records when necessary.
  \item \textbf{Record how every output was produced.}
  Each generated version records the Task Persona, Account Profile version, retrieved memories, derived style evidence, strategy versions, source materials, and, for a revision, the generation version from which it was created.
  \item \textbf{Choose the revision path according to the user's intent.}
  Local requests, such as changing the tone or title, continue through the Stage 3 revision process. Changes to the audience, topic, core problem, value proposition, content goal, or audience stage start a new Stage 1 clarification and Task Persona finalization cycle.
  \item \textbf{Run generation as recoverable background work.}
  Generation and revision are executed as asynchronous jobs that continue independently of the original HTTP request. Repeated equivalent requests do not create duplicate work, jobs are isolated between user accounts, and interrupted processing can be recovered or safely resumed.
\end{itemize}

\clearpage

\begin{figure*}[t]
  \centering
  \includegraphics[width=\textwidth]{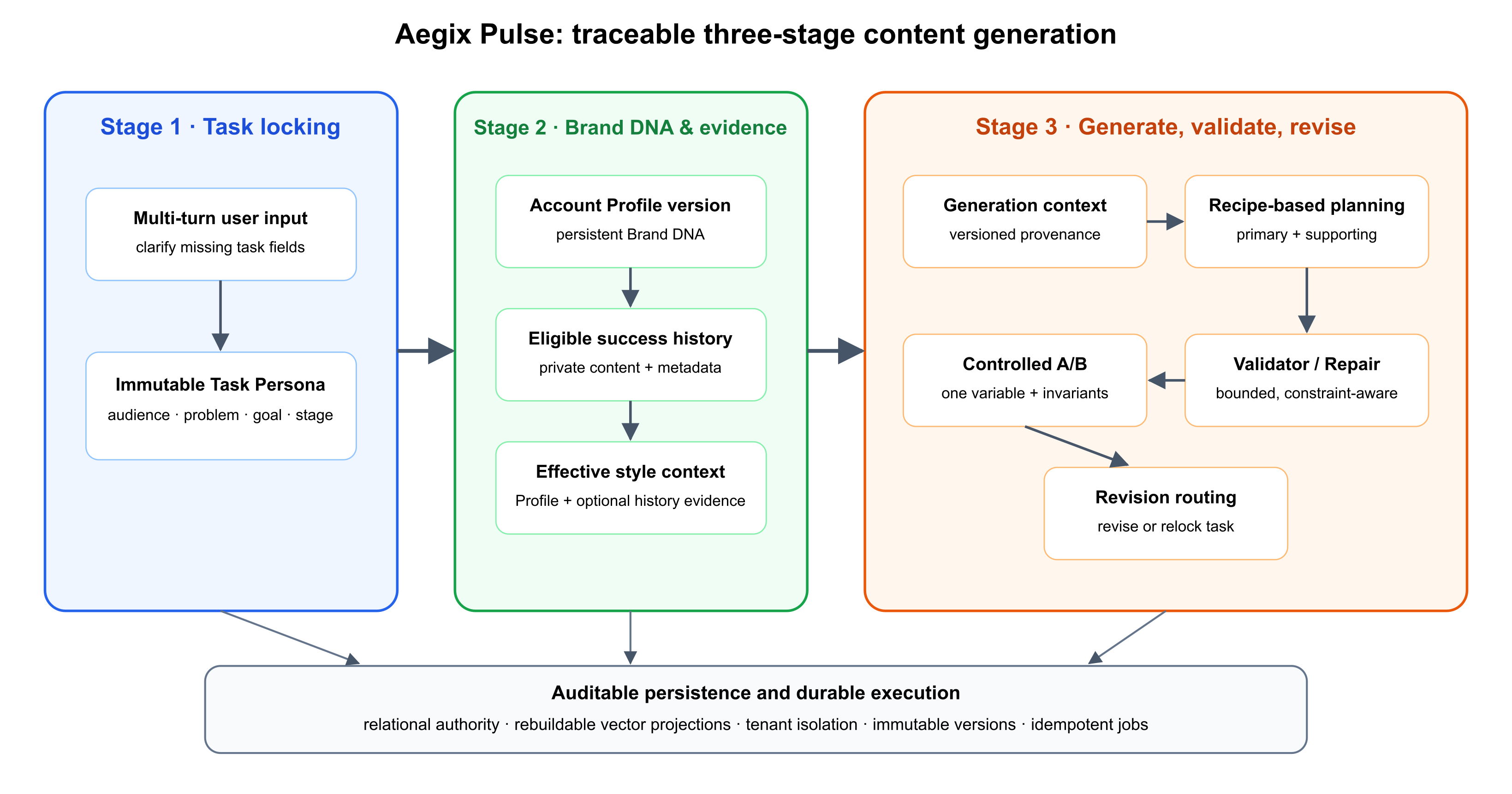}
  \caption{The three-stage \system{} architecture. Task-local state, persistent identity, performance-grounded memory, and revision lineage remain separately attributable.}
  \label{fig:architecture}
\end{figure*}

\subsection{Stage 1: dialogue and Task Persona finalization}

Stage 1 collects the user's requirements through a multi-turn conversation associated with a session identifier. From this conversation, the model constructs a Task Persona that describes the target audience, topic or product attributes, pain points, value proposition, content goal, and audience stage. If essential information is missing, the system asks a specific clarification question and continues the same session. Once the required information is complete and the model reaches the specified confidence threshold, the Task Persona is finalized and stored as the versioned task context for the subsequent stages.

Stage 1 also records which Account Profile version is active when the task is finalized. The Account Profile represents the user's long-term Brand DNA and remains separate from the Task Persona, which describes only the current content task. The finalized Task Persona is stored as a permanent relational record and is linked to the exact Account Profile version and source materials used for that task. A new content task starts with a new session. If the user later changes the Account Profile, the system creates a new Profile version instead of changing the Profile information associated with earlier generations.

\subsection{Stage 2: Account Profile and historical style evidence}

Stage 2 combines the active Account Profile (Brand DNA) with evidence from the user's own successful historical content. Historical content is considered only when enough performance data is available and the content meets a predefined, versioned success rule. In production, this rule can compare a post's performance with the user's historical baseline. If the user does not yet have enough historical data to support a reliable comparison, the system records the result as insufficient evidence rather than classifying the post as successful or unsuccessful.

Each retrieved item retains relevant metadata, such as its performance score, content goal, publication time, and generation identifier. This metadata supports ranking, filtering, and traceability, so the model does not receive historical text without information about its source and performance evidence.

The model summarizes reusable expression patterns from eligible historical content, including tone, pacing, rhetorical habits, and recurring presentation styles. The resulting successful-history style evidence is stored as a separate record linked to the current generation. It is not an additional Persona and does not change the Task Persona or the Account Profile (Brand DNA).

If no historical content meets the eligibility requirements, the style-evidence field remains null. In this case, the system uses the Account Profile as the primary source of style guidance and falls back to domain defaults only when the Profile does not provide enough information. When eligible historical evidence is available, the system combines it with the Account Profile in an Effective Style object. This object records the final style guidance, applicable constraints, and the strategy versions used to produce it.

\subsection{Stage 3: controlled generation and revision}

Stage 3 creates the Generation Context from the finalized Task Persona, the exact Account Profile version, retrieved private memories, successful-history style evidence, the Effective Style object, and relevant source materials. The system selects one primary Recipe to define the main content structure and may add supporting Recipes for specific structural requirements. The primary and supporting Recipes are recorded separately so that their contributions remain traceable.

For the initial generation, the system produces two candidate versions. Both candidates follow the same task, brand, factual, structural, and formatting requirements. They differ only in one clearly defined variable, such as the opening style or narrative approach. This design provides meaningful alternatives while keeping the comparison controlled.

Recipe-based planning has been implemented as part of Stage 3 and has passed the relevant engineering checks. However, the formal experiment in this paper does not evaluate whether Recipe-based planning improves content quality. Recipe planning is disabled in all E0--E3 and R0--R1 conditions, so none of the observed experimental effects are attributed to Recipe.

After generation, the Validator checks whether the output follows the brand requirements, stays within the available factual evidence, satisfies the selected Recipe, follows the platform's formatting rules, and preserves the requirements shared by the two controlled candidates. If the Validator identifies a violation, the Repair process can revise the output within a predefined maximum number of attempts. During repair, protected information---such as the task definition, verified facts, and content that was not requested to change---must remain unchanged.

Recipe, Validator, and Repair are implemented components of the production system and have passed their relevant engineering checks. However, they are disabled in the formal experiment reported in this paper. The experiment therefore does not claim that these components improve content quality relative to the comparison conditions.

When a user requests a revision, the system first determines whether the feedback asks for a local content change or changes the definition of the task itself. A local request---such as adjusting the title, tone, or wording---continues through the Stage 3 revision process. The revised content receives a new Generation ID and is linked to the exact previous version from which it was created.

Before revision, the system records which information must remain unchanged and which parts may be modified. Because the user is requesting a specific change to an existing version, the revision produces one updated candidate rather than an unnecessary A/B pair.

If the feedback changes the target audience, topic, core problem, value proposition, content goal, or audience stage, the backend returns the internal status \texttt{RELOCK\_REQUIRED}. This status indicates that the request changes the task definition and should begin a new Stage 1 clarification and Task Persona finalization cycle. The frontend starts the new session only after the user confirms this action.

\subsection{Persistence, tenancy, and durable jobs}

The production service provides authenticated API endpoints for content generation, revision, and Generation Job management. A relational database stores the authoritative records for users, Account Profile versions, finalized Task Personas, generated content, source-material relationships, revision history, background jobs, publication records, and performance snapshots.

A vector store supports semantic retrieval through rebuildable indexes. Separate indexes are used for Account Profile retrieval, user-private historical memories, and the shared Recipe library. Queries involving user-owned data are always restricted to the authenticated user or tenant. Shared Recipe retrieval is limited to the curated shared library and does not expose another user's private posts, Account Profile, performance data, or extracted style evidence.

Initial generation and revision are executed as asynchronous background jobs. When creating a job, the client provides an idempotency key. If the same request is submitted again with the same key, the backend returns the existing job instead of creating duplicate work. If the same key is reused for a different request, the backend rejects the request.

Only one job may be active for the same account, domain, and session at a time. However, a user can run multiple jobs in parallel when they belong to different sessions. Each job is stored persistently and moves through queued, running, and terminal states, allowing the frontend to poll for progress even after the original HTTP request has ended.

A successfully completed background job does not necessarily mean that content generation has been finalized. The business result is reported separately through internal statuses such as \texttt{CLARIFYING}, \texttt{LOCKED}, \texttt{REVISED}, and \texttt{RELOCK\_REQUIRED}. This distinction separates successful job execution from the outcome of the content workflow.

\subsection{Capability verification and evidence boundaries}

Table~\ref{tab:capabilities} distinguishes three questions for each system capability: whether it has been implemented, how its engineering behavior has been verified, and whether the experiment in this paper shows that it improves output quality. Engineering verification and statistical effectiveness require different types of evidence. A capability may be fully implemented and pass its engineering checks even when its effect on content quality has not yet been evaluated or confirmed statistically.

\begin{table*}[p]
\centering
\small
\caption{System capabilities, verification methods, and evidence supported by this study.}
\label{tab:capabilities}
\begin{tabularx}{\textwidth}{@{}p{0.16\textwidth}X X p{0.19\textwidth}@{}}
\toprule
Capability & Implemented mechanism & Evidence & Paper status \\
\midrule

Multi-turn clarification &
Session-based follow-up questions continue until the Task Persona can be
finalized. &
API and workflow tests; quality effect not tested separately. &
Engineering verified. \\

Task Persona &
Records the audience, topic, problem, value proposition, content goal,
and audience stage for one task. &
Contract tests; H1 showed a small, non-conclusive improvement. &
Engineering verified; small preliminary signal. \\

Account Profile (Brand DNA) &
Provides versioned long-term brand identity separately from the Task
Persona. &
Contract tests; H2 improved brand consistency by 0.1562 points, but was
not statistically conclusive. &
Engineering verified; promising preliminary evidence. \\

Successful-history style evidence &
Extracts reusable style patterns from eligible user-private historical
content. &
Contract tests; H3 showed no additional benefit beyond Account Profile
in this experiment. &
Engineering verified; no additional benefit observed. \\

Brand and factual constraints &
Validator detects violations and bounded Repair corrects eligible
problems. &
Validator and Repair tests; disabled in the formal experiment. &
Engineering verified; effectiveness not tested. \\

Recipe-based planning &
Uses one primary structural Recipe with separately recorded supporting
Recipes. &
Planning and persistence tests; disabled in E0--E3 and R0--R1. &
Engineering verified; effectiveness not tested. \\

Controlled A/B generation &
Produces two candidates that differ on one declared variable. &
Schema and invariant tests; user-choice benefit not tested. &
Engineering verified. \\

Context-preserving revision &
Revises locally while preserving task and brand context and recording
the exact parent version. &
Contract tests; H4 improved preservation by 0.2917 points, but was not
statistically conclusive. &
Engineering verified; promising preliminary evidence. \\

Task redefinition routing &
Starts a new Stage 1 cycle when feedback changes the task definition. &
Routing and API tests; user benefit not tested separately. &
Engineering verified. \\

Tenant and task isolation &
Restricts user data and active jobs by account and session. &
Authorization, cross-account 404, isolation, and concurrency tests. &
Verified within tested contracts. \\

Durable asynchronous jobs &
Supports idempotent creation, persistent state, polling, and stable
terminal results. &
Job lifecycle and API integration tests. &
Engineering verified. \\

Auditability &
Records versions, parent relationships, hashes, attempts, and frozen
experiment manifests. &
All 480 generation and 1,440 Judge records passed audit. &
System and research capability verified. \\

Production feedback model &
Records publications, performance snapshots, success decisions, and
memory eligibility. &
Schema and API tests; real business outcomes not evaluated. &
Implemented foundation. \\

\bottomrule
\end{tabularx}
\end{table*}

The table separates two types of evidence. Engineering tests show whether a capability has been implemented correctly and works according to its defined requirements. The formal experiment asks a different question: whether that capability improves content quality. Based on the engineering
evidence, we can state that \system{} supports traceable clarification, constraint checking, revision history, account isolation, and recoverable execution. However, these engineering results alone do not show that the system always produces better content or improves business outcomes. Such claims require separate controlled experiments and real production data.

\clearpage

\section{Methods}

\subsection{Research questions and hypotheses}

We preregistered four directional hypotheses:
\begin{itemize}
  \item \textbf{H1:} Adding the Task Persona improves task-intent alignment
  (E1 $>$ E0).
  
  \item \textbf{H2:} Adding the Account Profile improves brand consistency
  beyond the Task Persona alone (E2 $>$ E1).
  
  \item \textbf{H3:} Adding successful-history style evidence further
  improves brand consistency beyond the Task Persona and Account Profile
  (E3 $>$ E2).
  
  \item \textbf{H4:} Providing the original task and brand context during
  revision improves task preservation compared with plain revision
  (R1 $>$ R0).
\end{itemize}

These were the only confirmatory comparisons defined before the formal
experiment. The experiment did not evaluate Recipe-based planning,
Validator/Repair, automatic Task Persona extraction, retrieval accuracy,
or business outcomes.

\subsection{Experimental conditions}

The four initial-generation conditions used the same generator model, decoding settings, output schema, user task, and factual source material. They differed only in the additional context provided in each condition, as shown in Table~\ref{tab:conditions}.

\begin{table}[ht]
\centering
\small
\caption{Initial-generation ablation conditions.}
\label{tab:conditions}
\begin{tabular}{@{}lcccc@{}}
\toprule
Condition & Persona & Profile & History evidence & Validator \\
\midrule
E0 Plain LLM & No & No & No & No \\
E1 Task Persona & Yes & No & No & No \\
E2 Persona + Profile & Yes & Yes & No & No \\
E3 Persona + Profile + history evidence & Yes & Yes & Yes & No \\
\bottomrule
\end{tabular}
\end{table}

For reproducibility, the experiment database keeps the original E3 identifier, \path{E3_persona_profile_dna}. In this paper, we use the clearer term \emph{successful-history style evidence} for the additional E3 input. This input was extracted in advance from synthetic successful posts and was separate from the Account Profile (Brand DNA). The change in terminology does not affect the experiment, outputs, or statistical results.

The revision experiment included 48 selected tasks. R0 received the original E3 output and the user's revision request. R1 received the same output and request, together with the original Task Persona, Account Profile, successful-history style evidence, and generation context. Both conditions produced one revised version. Recipe and Validator/Repair were disabled in both conditions.

\subsection{Dataset construction and isolation}

The formal dataset contained 96 synthetic tasks on social media. It included 12 anonymous Account Profiles with eight tasks each, six content goals with 16 tasks each, four audience stages with 24 tasks each, and 12 business scenarios with eight tasks each. We selected 48 tasks for the revision experiment while keeping the Profiles, goals, audience stages, and scenarios balanced.

Each task included a user request, a manually checked Task Persona, an Account Profile, factual and wording constraints, source material or a clear marker that no source was provided, synthetic successful posts, style evidence extracted from those posts, and a local revision request. The synthetic successful posts were labelled \texttt{experimental\_verified\_success} and were kept separate from production memory and performance data.

We used 20 Pilot tasks and 24 previously reviewed development tasks only to find and diagnose problems. They were excluded from the formal results. Before running the models, we locked the 96-task formal dataset and recorded its SHA-256 hash. This hash makes it possible to verify that the dataset was not changed after the experiment began.

\subsection{Models and frozen prompts}

Generation used \texttt{deepseek-ai/DeepSeek-V3} through SiliconFlow’s OpenAI-compatible endpoint. Blind evaluation used a separate model family, \texttt{Qwen/Qwen3.5-122B-A10B}. The production system uses \texttt{BAAI/bge-m3} for embedding-related components, but the experiment used fixed condition inputs and did not evaluate retrieval performance. The provider did not provide an immutable revision identifier for the hosted model.

The scoped generator, revision strategy, Judge prompt, schema, aggregation policy, decoding settings, and randomization seeds were all versioned in the experiment manifest. Candidate order was reproducibly randomized within each task and replication. The Judge was not given group labels, generation prompts, model identity, repair counts, or system logs.

\subsection{Outcomes}

The primary outcomes were: (1) task-intent alignment, which measured whether the output matched the intended audience, problem, topic, value proposition, goal, and audience stage; (2) brand consistency, which measured alignment with the Account Profile, brand positioning, tone, wording rules, and expression boundaries; and (3) revision preservation, which measured whether the revised output followed the requested change while preserving the original task, unchanged content, facts, and brand style. Each outcome was rated on a 1--5 scale.

Revision preservation was assessed across five dimensions: preservation of the task definition, preservation of content that the user did not ask to change, completion of the requested change, preservation of facts, and preservation of brand style and content structure. We also recorded rule-based and operational measurements, including factual and constraint violations, title length, SEO format, prohibited expressions, generation success, latency, retries, token usage, and estimated cost. These measurements were not part of the 1--5 Judge scores or the confirmatory hypothesis tests.

\subsection{Automated and LLM evaluation}

Rule-based checks verified the output schema, title length, SEO-tag format, prohibited expressions, numerical claims, the single-candidate structure of revisions, and the version history of generated content. Factual accuracy and compliance with semantic constraints were evaluated by an LLM Judge using scoring rules fixed before the formal experiment. The Judge was not shown which experimental condition produced each output.

Each unique candidate-and-context pair was evaluated independently three times by the Judge. Identical pairs reused the same recorded evaluation to avoid introducing artificial differences. For initial generation, we took the median of the three scores for each candidate and then averaged the scores of candidates A and B within each task and condition. For revision, we first took the median score for each of the five dimensions and then used the lowest dimension score as the overall preservation score, as defined in advance. A dimension was sent for human review when its three scores differed by at least two points or when no score appeared more than once.

\subsection{Human review and adjudication}

Two reviewers independently evaluated a balanced subset of 24 tasks covering all Account Profiles, content goals, audience stages, and experimental conditions. They reviewed 96 initial-generation records and 48 revision records without knowing which condition produced each output. A score difference of two or more points was treated as a major disagreement. Disagreements about whether a violation occurred were also sent for third-person review.

Across the 144 reviewed records and their multiple evaluation dimensions, the two reviewers produced 129 dimension-level score disagreements of two or more points and 120 item-level disagreements about whether a violation occurred. Because one candidate could contain more than one disagreement, these counts do not represent 249 separate outputs. After duplicate candidates were removed, a third reviewer independently assessed 201 disputed candidates without seeing the earlier human scores or the LLM-Judge results. Following the rules set before the review, the third reviewer's score was used for each major dimension-level score disagreement; otherwise, the average of the first two reviewers' scores was retained. The third reviewer also made the final decision for each disputed violation. We reported the original agreement between the first two reviewers separately.

\subsection{Statistical analysis}

Each task under each experimental condition was treated as one unit of analysis, and conditions were compared within the same task. We report the mean, standard deviation, median, quartiles, and 95\% confidence intervals. The confidence intervals were calculated using 10,000 bootstrap resamples grouped by task. For the primary outcomes, we used two-sided paired permutation tests and paired Cohen's $d_z$ to measure effect size. Wilcoxon signed-rank tests were also used to check whether the conclusions remained similar under a different statistical method. Because H1--H4 were tested together, their $p$-values were adjusted using Holm's method with $\alpha=.05$. The adjudicated human review of 24 tasks was used only as descriptive external validation, not as a second confirmatory test. After the condition labels were revealed, the formal dataset, prompts, system, scoring rubric, exclusion rules, and analysis methods could no longer be changed.

\section{Results}

\subsection{Integrity and execution}

All 384 initial-generation runs and 96 revision runs completed successfully, producing 480 records in total. These records contained 768 initial A/B candidates and 96 single-candidate revisions. The audit confirmed that the output hashes, data formats, experimental-condition assignments, and revision version history were correct. All 1,440 planned Judge evaluations were complete, unique, and valid. All 864 candidates passed the SEO-format and prohibited-expression checks, while 863 passed the title-length check. The single title-length failure was kept in the analysis.

\subsection{Preregistered comparisons}

\begin{table}[ht]
\centering
\small
\caption{Results of the four preregistered comparisons. Mean differences are calculated as the experimental condition minus the comparison condition on a five-point scale. Each 95\% confidence interval applies to one comparison and is not adjusted for testing multiple hypotheses.}
\label{tab:results}
\resizebox{\textwidth}{!}{%
\begin{tabular}{@{}cllrrrrrl@{}}
\toprule
Hyp. & Comparison & Outcome & $n$ & Mean diff. & 95\% CI & Raw $p$ & Holm $p$ & Decision \\
\midrule
H1 & E1--E0 & Task intent & 96 & 0.0625 & [$-0.0573$, 0.1823] & .364196 & .728392 & Inconclusive \\
H2 & E2--E1 & Brand consistency & 96 & 0.1562 & [0.0208, 0.2917] & .030590 & .122360 & Positive; preliminary \\
H3 & E3--E2 & Brand consistency & 96 & $-0.0573$ & [$-0.2240$, 0.1042] & .540725 & .728392 & Not confirmed \\
H4 & R1--R0 & Revision preservation & 48 & 0.2917 & [0.0208, 0.5833] & .081055 & .243165 & Positive; preliminary \\
\bottomrule
\end{tabular}%
}
\end{table}

\begin{figure}[ht]
  \centering
  \includegraphics[width=\linewidth]{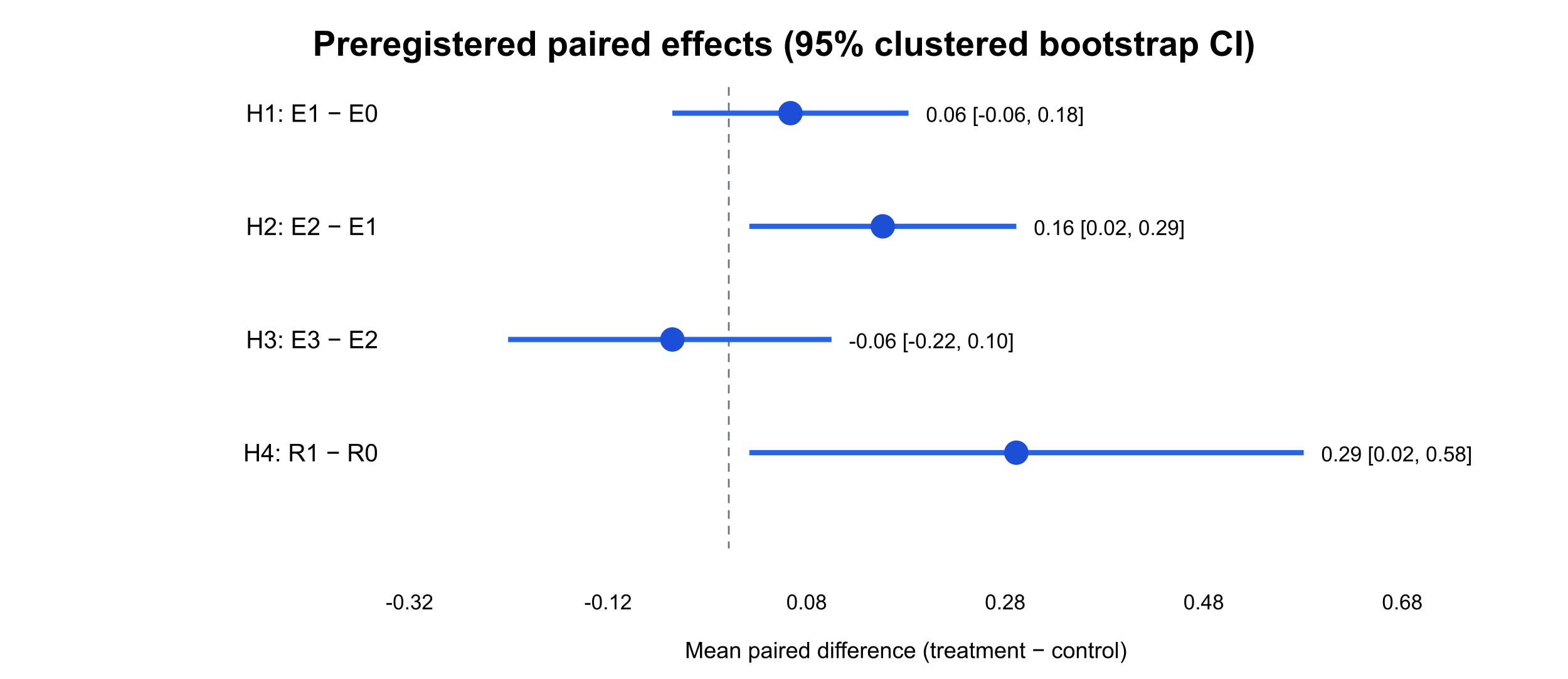}
  \caption{Results of the preregistered paired comparisons. Positive values indicate better performance after adding the component being tested.}
  \label{fig:effects}
\end{figure}

The Task Persona produced a small improvement in task alignment, but the evidence was not conclusive. Adding the Account Profile produced the largest improvement among the initial-generation comparisons. Its raw $p$ value was below .05, but it did not reach the prespecified significance threshold after correction for multiple comparisons. Adding successful-history style evidence produced a small negative difference in this experiment. Providing the original task and brand context during revision improved task preservation, but this result also remained preliminary after correction (Table~\ref{tab:results} and Figure~\ref{fig:effects}).

\subsection{Descriptive condition scores}

\begin{figure}[H]
  \centering
  \includegraphics[width=\linewidth]{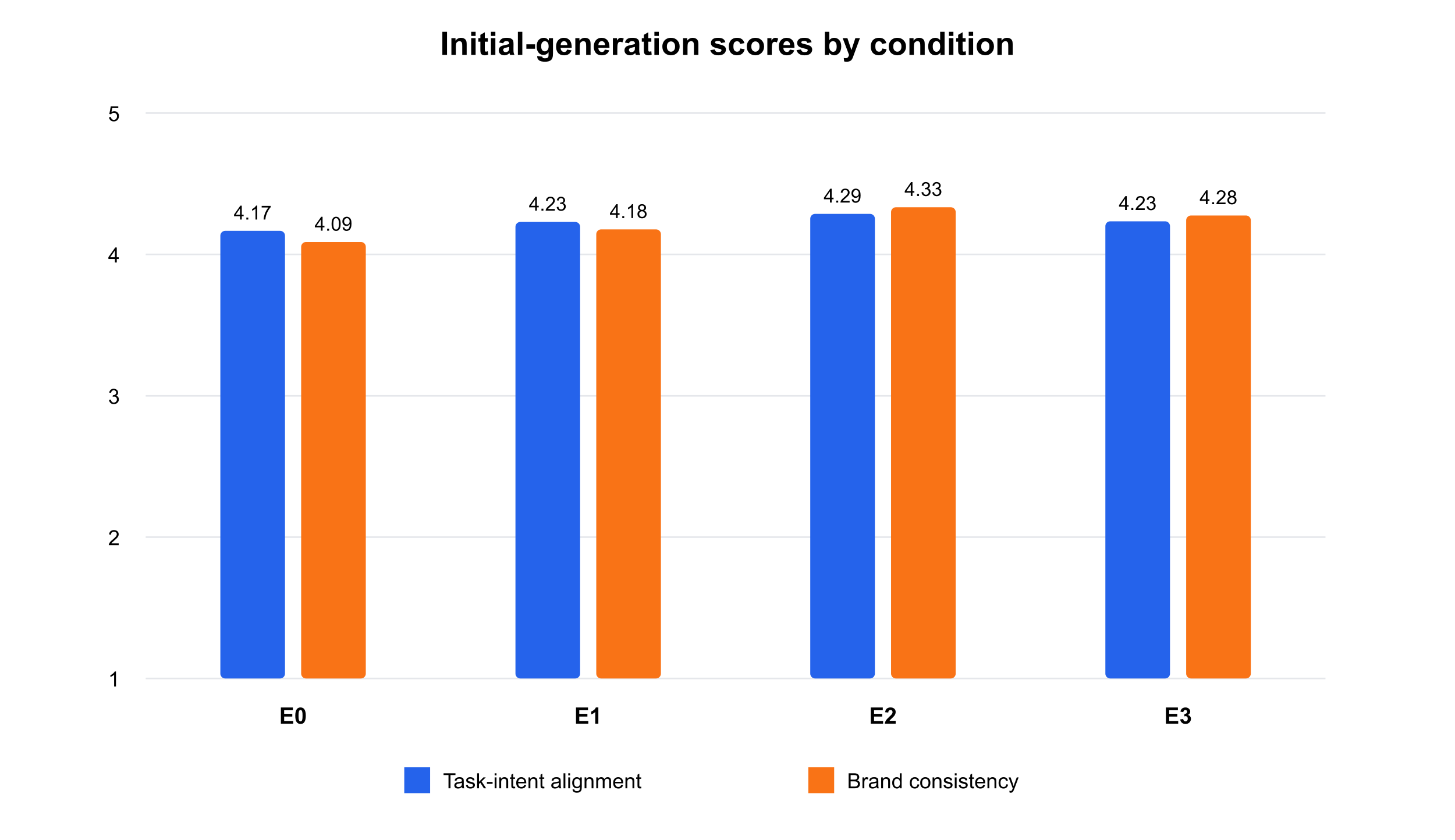}
  \caption{Mean task-intent and brand-consistency scores for the four initial-generation conditions (E0--E3).}
  \label{fig:initial}
\end{figure}

\begin{figure}[H]
  \centering
  \includegraphics[width=0.82\textwidth]{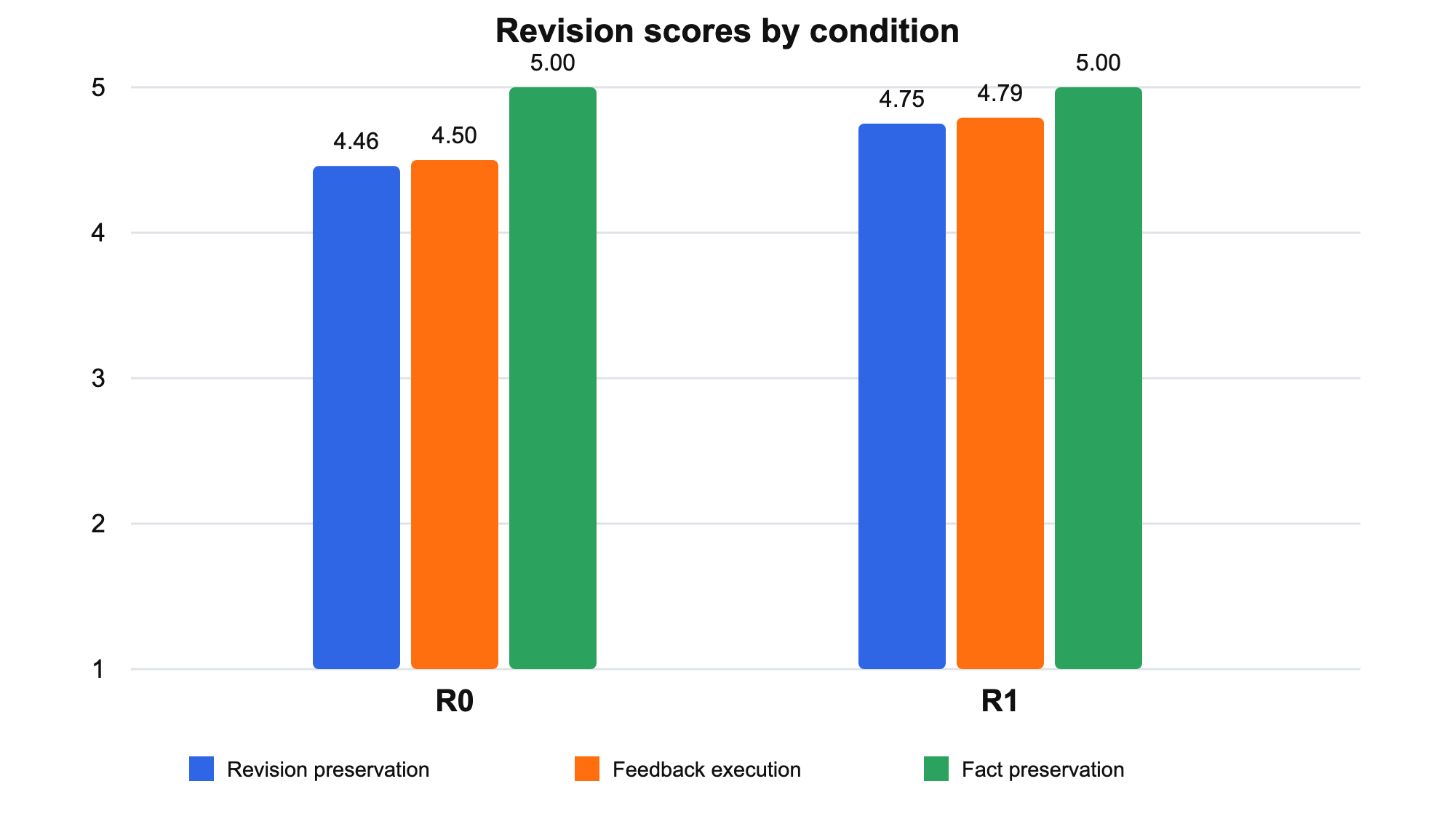}
  \caption{Mean scores for the two revision conditions. R0 performed plain revision, while R1 received the original task and brand context. The values are descriptive; the preregistered paired comparison is reported separately in Table~\ref{tab:results}.}
  \label{fig:revision-scores}
\end{figure}

The mean task-intent scores for E0--E3 were 4.1667, 4.2292, 4.2865, and 4.2344, respectively. Their mean brand-consistency scores were 4.0885, 4.1771, 4.3333, and 4.2760 (Figure~\ref{fig:initial}). Because the baseline scores were already high and many scores were close to the top of the five-point scale, there was limited room for further improvement. Mean revision-preservation scores were 4.4583 for R0 and 4.7500 for R1. Feedback-execution scores were 4.5000 and 4.7917, respectively, while both conditions received a fact-preservation score of 5.0000 (Figure~\ref{fig:revision-scores}).

\subsection{Judge stability and human validation}

Across the three LLM evaluations, 99.09\% of task-intent scores, 97.27\% of brand-consistency scores, and 97.92\% of feedback-execution scores differed by no more than one point. The corresponding quadratic-weighted agreement scores were 0.7093, 0.7768, and 0.8220, showing that the repeated LLM evaluations were generally consistent.

\begin{figure}[ht]
  \centering
  \includegraphics[width=\linewidth]{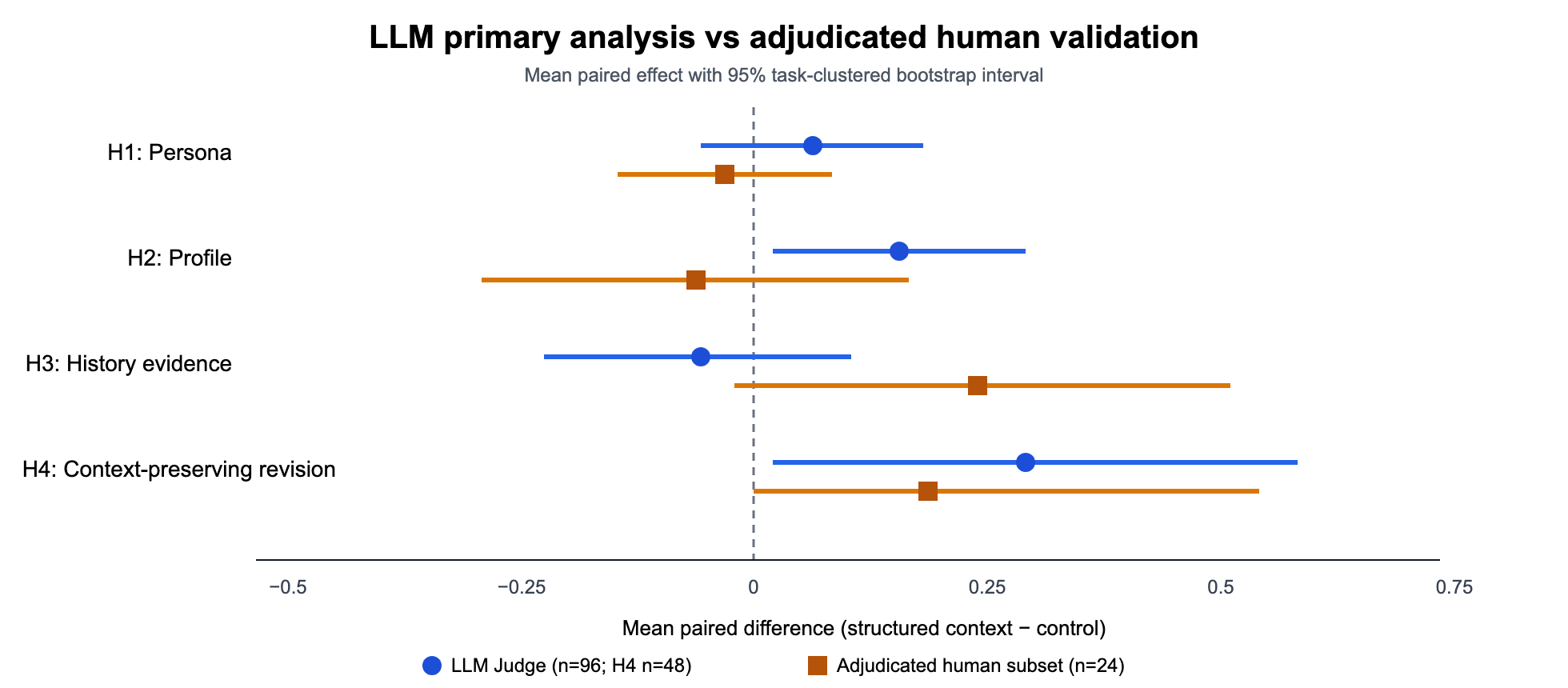}
  \caption{Full-set LLM effects and adjudicated-human subset effects.}
  \label{fig:human}
\end{figure}

Agreement between the two human reviewers was poor, with a quadratic-weighted kappa of 0 for both task intent and brand consistency. A third reviewer resolved all 129 major score disagreements and 120 disagreements about violations, but this did not change the original agreement estimate between the first two reviewers. After adjudication, the mean paired effects in the human-reviewed subset were $-0.0312$ for H1 (95\% CI [$-0.1458$, 0.0833]), $-0.0625$ for H2 ([$-0.2917$, 0.1667]), 0.2396 for H3 ([$-0.0208$, 0.5104]), and 0.1875 for H4 ([0, 0.5417]). Only H4 showed the same direction of change in both the human and LLM evaluations. Among the 240 candidate records reviewed by humans, 63 (26.25\%) were judged after adjudication to contain at least one factual or constraint-related issue (Figure~\ref{fig:human}).

\subsection{Operational metrics}

The estimated total model cost was CNY~28.508254, including CNY~1.776658 for content generation and CNY~26.731596 for evaluation. Judge calls took a median of 8,273.5 ms and an average of 8,884.12 ms. The system recovered from 19 rate-limit events and nine incomplete responses by recording each retry separately and resuming only the affected work. No formal experimental record was deleted or overwritten.

\section{Discussion}

\subsection{Empirical interpretation}

The experiment did not provide enough evidence to conclude that adding more structured context consistently improves content quality. After correction for multiple comparisons, none of the four hypotheses reached the prespecified significance threshold, and the measured effects were small. The clearest positive results came from adding the Account Profile and preserving the original context during revision. Both results support further study and provide useful directions for system development, but the current evidence is not yet sufficient to confirm that either approach is superior.

A Task Persona may provide limited additional value when the user's original request is already detailed. The Account Profile provides different information, including the brand's long-term tone, positioning, and wording rules, which may not appear in the current request. This difference may help explain the positive result for E2 compared with E1. However, the result should be tested again on a new dataset fixed before the experiment before it can be considered reliable.

Successful-history style evidence did not further improve brand consistency beyond the Task Persona and Account Profile in this experiment. One possible reason is that it repeated information already contained in the Account Profile. It may also have conflicted with the current task or provided weak guidance because it was extracted from synthetic history. These findings suggest that historical style evidence should not automatically take priority over the current task and Account Profile. They do not show that performance-based memory is ineffective. Its value may depend on real long-term performance data, retrieval accuracy, the amount of available history, and the content domain.

The positive revision result suggests that providing the original task and brand context can help preserve important information when users request small changes. This supports the system's separation between local Stage 3 revision and starting a new Stage 1 task when the task definition changes. However, R0 also received high scores, and the improvement did not reach the prespecified significance threshold after correction. Future studies should evaluate more difficult revision requests with conflict levels defined before the experiment, rather than adjusting the system based on the current test cases.

The mean score for every initial-generation condition was above 4.0, leaving limited room for improvement on a five-point scale. Using a capable current model as the baseline gives a more realistic comparison than deliberately choosing a weak baseline, but it also makes small improvements more difficult to detect. Future studies may need more challenging tasks and more sensitive evaluation methods, which should be designed independently of the results reported here.

The repeated LLM evaluations were consistent, but the human-reviewed subset did not show the same direction of change for most effects. At the same time, the two original human reviewers often disagreed with each other. Therefore, the results do not show that LLM evaluation is equivalent to human judgment, or that human judgment is automatically more reliable. They show only that the fixed LLM evaluator produced repeatable scores, while agreement with independent human evaluation remained limited and depended on the reviewers and evaluation method.

\system{}'s main contribution in this first paper is its system architecture and evaluation method. The system keeps different types of context separate, records where each input and output came from, tracks the full revision history, and supports reliable production execution. The evaluation framework tests these components separately, defines the research claims before the experiment, resumes interrupted runs without deleting earlier records, and keeps both positive and negative results. Together, these features provide a reliable foundation for future studies, even though the improvements measured in this first experiment were modest.

\subsection{Engineering validity is not outcome effectiveness}

Table~\ref{tab:capabilities} is important for understanding the limits of this paper's claims. \system{} provides multi-turn clarification, immutable Task Persona records, account isolation, revision history, constraint reports, idempotent jobs, and an audit trail. These features have clear engineering requirements that can be tested directly. They make workflow changes controllable, interrupted work recoverable, access restrictions enforceable, and generated outputs traceable to the exact input and component versions used. These production capabilities address risks that are not captured by a one-step prompt comparison, even when the two approaches receive similar content-quality scores.

However, engineering tests and effectiveness experiments answer different questions. Engineering tests ask whether the system works according to its requirements, while effectiveness experiments ask whether a specific component improves a measured result compared with a baseline. This study evaluates content-quality effects only for the Task Persona, Account Profile, successful-history style evidence, and the use of original context during revision. It does not measure whether clarification, Recipe, Validator/Repair, asynchronous jobs, account isolation, or auditability improves content quality. We therefore present these capabilities as implemented and engineering-verified system contributions, without claiming that their effects on content quality have been established.

This distinction also makes the findings easier to interpret from a product perspective. The lack of an additional quality improvement from successful-history style evidence does not mean that source tracking, privacy protection, version control, or safe fallback behavior has no value. At the same time, successful database and API tests cannot prove that historical style evidence improves content quality. Each conclusion must therefore be limited to what its supporting evidence can actually show.

\section{Limitations}

The experiment used synthetic tasks, Account Profiles, and successful posts, and focused only on Chinese social media content for cross-border e-commerce sellers. We do not yet know whether the findings apply to real users, other platforms, other languages, or brands that change over time. The study evaluated content quality offline and did not measure exposure, engagement, conversion, revenue, user retention, or satisfaction. It also did not use real or simulated Performance Snapshot data as an experimental outcome, so this paper makes no claims about business-performance improvements.

The formal content-quality evaluation used one hosted LLM Judge model. Repeating each evaluation improved consistency but could not remove possible systematic bias. The provider also did not specify a fixed model revision, so future provider-side updates may affect exact reproduction of the results. Agreement between the two original human reviewers was poor. Although a third reviewer resolved the disagreements, this did not improve the reliability of the original ratings. The human evaluation covered only 24 tasks and was used to describe and cross-check the results, not to confirm the hypotheses statistically.

Most scores were close to the top of the five-point scale, leaving limited room to detect improvement. The small number of score levels may also hide minor differences between conditions. In the revision experiment, the overall score was the lowest score among five dimensions, so one weak dimension could substantially reduce the final result.

The formal experiment used Task Personas, Account Profiles, and successful-history style evidence that had been checked in advance. It did not test how accurately the production system automatically extracts tasks, retrieves memories, identifies successful content, or extracts style patterns from historical posts. Results from the complete production workflow may therefore differ from the findings reported here. Recipe and Validator/Repair were disabled in the formal experimental conditions. Their implementation has been verified through engineering tests, but this paper does not show that they improve content quality.

Finally, the formal dataset has now been examined and its results are known. It cannot be used to adjust the system, LLM Judge, scoring rules, metrics, or exclusion criteria and then be reused as independent evidence. Any improved version must be evaluated under a new preregistered protocol using a separate dataset.

\section{Future Work}
The next study should test the Account Profile and context-preserving revision on a new dataset created before any outputs are examined. It should include more difficult and conflicting task requirements and use human reviewers who receive clearer guidance and calibration. A later production study can examine whether style evidence extracted from real successful posts improves brand consistency and user acceptance over time. That study must clearly separate simple correlations in historical performance from improvements actually caused by the system component being tested.

Future work will continue to improve and independently evaluate Recipe-based structural planning in Stage 3. A new preregistered experiment should compare generation with and without Recipe to determine whether reusable content structures improve structural fit, compliance with requirements, and stability during revision. Validator effectiveness should be tested in a separate experiment so that its contribution is not mixed with other system components. The automatic processing used in Stage 1 and Stage 2, including task extraction and historical-evidence retrieval, should also be evaluated using separately annotated datasets.

Finally, future evaluations should combine rule-based checks, several independent LLM Judge models, trained and calibrated human reviewers, and outcomes that reflect actual user experience. When these evaluation methods disagree, the differences should be reported and analyzed directly rather than merged into a single score and treated as an objective result.

\section{Conclusion}

\system{} shows how a production content-generation system can keep the current task, long-term Brand DNA, style evidence from successful historical content, controlled generation, and revision history separate and traceable. In the preregistered component-level experiment, adding the Account Profile and preserving the original context during revision both produced encouraging improvements. However, the current evidence did not reach the prespecified significance threshold after correction for multiple comparisons. The Task Persona alone produced only a small improvement, while successful-history style evidence did not provide an additional LLM-rated benefit beyond the Task Persona and Account Profile in this experiment. The human review also showed that conclusions can vary depending on the evaluation method and evaluator.

The findings do not show that adding more personalization context always produces better results. Instead, they support an architecture that gives each context layer a clear purpose, defines which source takes priority, and records where the information came from. They also support an evaluation process that preserves both positive and negative findings. Together, these practices provide a reliable foundation for future production studies and new preregistered experiments using independent datasets.

\section*{Ethics, Data Governance, and Reproducibility}

The study used synthetic tasks, Account Profiles, and successful histories. It did not use ordinary production-user content, private messages, real platform-performance data, or simulated performance data presented as real evidence. Experimental records were kept separate from production memory, human-review files contained no account credentials, and generated content was not automatically published.

We retained the frozen protocol, dataset hash, experimental definitions, Judge prompt hash, analysis scripts, audit results, and adjudication records for internal reproducibility. Public releases will exclude provider identifiers and any material that the company is not permitted to distribute. Future studies involving real users or performance data will require explicit authorization, data minimization, access isolation, retention rules, and appropriate ethical review. Because the hosted model provider does not expose a fixed model revision, exact future reproduction may be affected by provider-side updates.

\section*{Competing Interests}

The authors are affiliated with Aegix Insight, which develops \system{}, and therefore have a commercial interest in the system. To reduce potential bias, the experimental protocol and analysis plan were fixed before the formal results were examined, and results that did not support the hypotheses are reported openly.

\bibliographystyle{plainnat}
\bibliography{references}

\end{document}